\documentclass[runningheads]{llncs}
\usepackage{cite}
\usepackage{amsmath,amssymb,amsfonts}
\usepackage{blkarray}
\usepackage{subcaption}
\usepackage{algorithmic}
\usepackage{graphicx}
\usepackage{textcomp}
\usepackage{hyperref}
\usepackage{multirow}
\usepackage{xcolor}

\hypersetup{
    colorlinks,
    linkcolor={red!50!black},
    citecolor={blue!50!black},
    urlcolor={blue!80!black}
}

\begin{document}

\title{Alignment-based conformance checking over probabilistic events\\}
%
%
\author{Jiawei Zheng\inst{1}\orcidID{0000-0002-6515-6423} \and
Petros Papapanagiotou\inst{1}\orcidID{0000-0003-0928-6108} \and Jacques D. Fleuriot\inst{1}\orcidID{0000-0002-6867-9836} }

\authorrunning{J. Zheng et al.}
%
\institute{School of Informatics, University of Edinburgh, Edinburgh, UK\\
\email{\{jw.zheng,pe.p,jdf\}@ed.ac.uk}\\ 
}

\maketitle
\begin{abstract}%
Conformance checking techniques allow us to evaluate how well some exhibited behaviour, represented by a trace of monitored events, conforms to a specified process model. Modern monitoring and activity recognition technologies, such as those relying on sensors, the IoT, statistics and AI, can produce a wealth of relevant event data. However, this data is typically characterised by noise and uncertainty, in contrast to the assumption of a deterministic event log required by conformance checking algorithms. In this paper, we extend alignment-based conformance checking to function under a probabilistic event log. We introduce a weighted trace model and weighted alignment cost function, and a custom threshold parameter that controls the level of confidence on the event data vs.\ the process model. The resulting algorithm considers activities of lower but sufficiently high probability that better align with the process model. We explain the algorithm and its motivation both from formal and intuitive perspectives, and demonstrate its functionality in comparison with deterministic alignment using real-life datasets.

\end{abstract}

\section{Introduction}\label{sec:Introduction}

Conformance checking is the task of comparing the behaviour captured by event logs that have been recorded during the execution of a process to the intended behaviour of a corresponding process model~\cite{carmonaConformanceCheckingRelating2018}. It evaluates how well the execution matches the modelled process and allows us to detect when things are diverging from a desired or expected workflow. 
A widely used conformance checking approach is based on the \textit{alignment} of the events in the log with the activities in the process model, where any misalignments are potential deviations~\cite{adriansyahConformanceCheckingUsing2011}. 

With the advances of sensor technologies, the Internet of Things (IoT) and modern Artificial Intelligence, we are now generating an abundance of data that can be used to detect events and processes as they are being performed in the physical world. However, this data is typically noisy, with errors caused by various factors such as the environment, the accuracy of the sensors, faults, malfunctions, and the inherent uncertainty of statistical and machine learning techniques used to analyse the data. The standard \textit{alignment} conformance checking approach is only able to fit deterministic events that are assumed to reflect reality with certainty~\cite{adriansyahConformanceCheckingUsing2011,IzackUncertainChallenge2021}. This means that probabilistic event data has to be reduced to a deterministic event log in order to be used. This removes information from the event log, thus lowering the confidence on any detected deviations. 

We present an extension of alignment-based conformance checking that works under the assumption of a probabilistic event log with a categorical distribution over a set of activities. We introduce a cost function that takes activity probabilities into consideration, and a custom parameter that allows the algorithm to align activities of lower but sufficiently high probability that better agree with the process model, as opposed to always assuming the most-probable activities occurred. In effect, this leverages the knowledge captured in the process model to address levels of uncertainty in the event data, with the aim of reducing the number of false positives and false negatives in deviation detection. We further motivate our work with an example from recent research in Activities of Daily Living (Section~\ref{sec:motivation}). We give a formal presentation of our algorithm (Section~\ref{sec:algorithm}), explain the operation and intuition behind its parameter (Section~\ref{sec:threshold}), and evaluate the performance on real-life datasets (Section~\ref{sec:experiment}). We implement our algorithm as a Python package based on the PM4Py framework~\cite{PM4PYpaper2019} and available at~\url{https://github.com/jia-wei-zheng/alignment-over-probabilistic-events}.

\section{Motivating Example}
\label{sec:motivation}
Smart home monitoring of Activities of Daily Living (ADLs) has seen increased attention in recent research, particularly in the context of home care~\cite{zhengPredictiveBehaviouralMonitoring2022}. 
For instance, ADL data can be used to analyse the daily routine of an older adult (as a personal process model) and detect when they deviate from that healthy behaviour using conformance checking~\cite{behavealwayssame,zhengPredictiveBehaviouralMonitoring2022}, such as skipping medication, etc. 

ADL data is typically based on smart home and wearable sensors. A Human Activity Recognition (HAR) algorithm, often based on supervised learning, helps classify that data into distinct daily activities, such as sleeping, eating, reading, etc. Subsequent analysis, including for conformance, usually assumes the most probable activity is the one that took place, and the alternatives, no matter how likely, are never considered~\cite{IzackUncertainChallenge2021,sztylerSelftrackingReloadedApplying2016}. However, the result of HAR is uncertain, usually taking the form of a categorical probability distribution over the possible classes of activities, with some classes being particularly difficult to distinguish. 
For example, given some sensor input, a HAR algorithm output may indicate there is 33\% probability that \textit{drinking from cup} took place, 34\% probability that \textit{answering phone} happened, and 33\% probability it was actually \textit{spraying from a bottle}, all three of which options have a similar pattern in the sensor data~\cite{guptaObjectsActionApproach2007}. 

In the context of ADL monitoring for care, a misclassification of \textit{drinking from cup} as \textit{answering phone}, which is only 1\% more likely in our example, may lead to an alert of a deviation, for instance related to the levels of hydration of the person. This would be an unnecessary, false positive alert, caused by the fact that the conformance analysis only considers the most probable activities. Instead, we want an algorithm that operates under an assumption of uncertainty, such that it considers the process model as an additional source of information of what may have occurred in reality, beyond the noisy sensor and HAR data.

In contrast, if the probability of an activity is too low (e.g.\ a 3\% probability of \textit{drinking from cup} compared to a 64\% probability of \textit{answering phone}), we may want to consider it as non-conforming, even if it is expected by the process model. 
Therefore, incorporating the level of \emph{confidence} in the pattern of the process model compared to the uncertainty of the event log is a key requirement.

The same challenges are emerging in other domains in which sensors and machine learning are featured prominently, such as smart manufacturing (the so-called \emph{Industry 4.0}), autonomous vehicles and healthcare, which require conformance to production flows, driving regulations and care pathways respectively.

\section{Related work}\label{sec:related work}

Alignment conformance checking algorithms transforms the problem of computing alignments between event logs and process models into an optimal search problem using the A* algorithm, i.e., it minimises the total alignment cost~\cite{adriansyahConformanceCheckingUsing2011}.

Recent work on conformance checking under uncertainty has three focal points: missing data~\cite{felliConformanceCheckingUncertainty2022,pegoraroConformanceCheckingUncertain2021},  uncertain event-activity mappings~\cite{vanderaaEfficientProcessConformance2020}, and uncertain process models~\cite{bergamiProbabilisticTraceAlignment2021}. Koorneef et al.\ present a probability-based alignment cost function, based in frequencies of events in historical event logs, to obtain the most \emph{probable} alignment~\cite{koorneefAutomaticRootCause2018}. We adapt this in the context of uncertain events. 

Even more recently, Cohen et al.\ presented a review of the challenges associated with using probabilistic event data in conformance checking techniques~\cite{IzackUncertainChallenge2021}. They highlighted the demand for new conformance checking techniques that can adapt multiple types of uncertain event data, which further motivates our work.

Based on the challenges identified by Cohen et al., Bogdanov et al.\ presented a conformance checking algorithm over stochastic logs, which takes the probability of event occurrences into account explicitly, similarly to our algorithm~\cite{bogdanovConformanceCheckingStochastically2022a}. They apply their algorithm on \textit{trace recovery}, which represents recovering the original trace from a probabilistic one based on a maximal alignment to a reference process model~\cite{bogdanovTraceRecoveryStochastically2022}. 
However, in their approach an event is aligned to a corresponding activity in the process model regardless of the probability that activity actually occurred, as long as it is non-zero. This assumes the occurrence of activities with very low or even close to zero probability, as long as they align with the process model, therefore discarding potential true positive deviations.

Overall, the need for conformance checking under uncertainty has been recognised by researchers in the area, but the related research is still in its infancy.

\section{Preliminaries}\label{sec:pre}

Our approach is built upon well-established literature on alignment-based conformance checking~\cite{adriansyahConformanceCheckingUsing2011, carmonaConformanceCheckingRelating2018}. We provide brief definitions of relevant concepts next so that this paper is self-contained. 

\begin{definition}[Petri net]
Given a set of activities $A$, a \textit{Petri net} is defined as a tuple $N = (P,T,F,\alpha)$, where $P$ and $T$ are sets of places and transitions respectively. $F$ is a set of arcs representing flow relations between transitions and places, so that $F \subseteq (P \times T)\cup(T \times P)$. A labelling function $\alpha : T \rightarrow A\cup \{\tau\}$ assigns either an activity from $A$ or $\tau$ (immediate transition not associated with any activity) to each transition in $T$. 
\end{definition}

\begin{figure}[htbp]
    \centering%
    \includegraphics[width=.55\columnwidth]{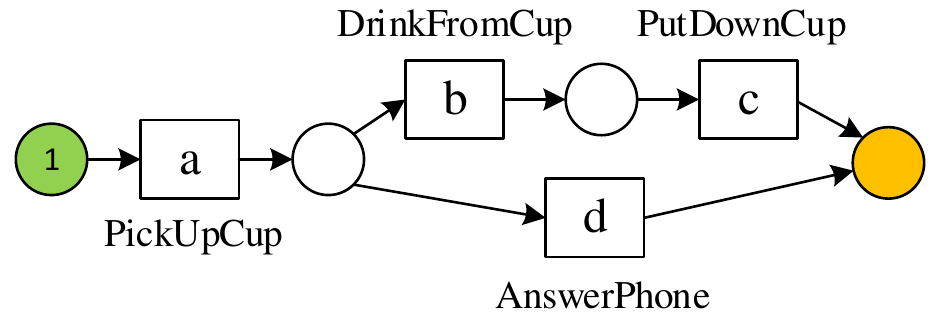}
    \caption{Example of a process model.}%
    \label{fig:process model}%
\end{figure}

Petri Nets offer a standard representation for \emph{process models}.
Fig.~\ref{fig:process model} shows an example Petri Net of a process model.

An \textit{event log} captures information about the execution of multiple cases (or instances) of a process model. Each case is represented by a trace of events $ \sigma = \langle e_{0}, e_{1}, ..., e_{n}\rangle$ that correspond to \emph{observations} of the activities that occurred during the process. We assume each event corresponds to a single activity in the process. An event is \emph{deterministic} if it can be mapped to its activity in a deterministic way. 

When aligning a trace of events to a model, the trace is transformed into a \textit{Petri net} called a \textit{trace model}.

\begin{definition}[Trace model]\label{def:trace model}
Given a set of activities $A$, a sequence of events over these activities $\sigma = \langle e_{0}, e_{1}, ..., e_{n}\rangle \in A^{*}$ can be converted into a linear \textit{Petri net} $N_{t} = (P^{t},T^{t},F^{t},  \alpha^{t})$, which we call a \emph{trace model}. In this, each event $e_i$ is mapped to a transition in $T^t$. The transitions are then interleaved with places in $P^t$ to form a linear sequence.
\end{definition}

The trace model can be combined with a given process model in a single Petri Net called a \emph{synchronous product net}. In this, each pair of transitions from the two models labelled with the same activity are represented by a \textit{synchronous move} $T^{SM}$, e.g., $(a,a)$ for activity $a$, meaning that an event corresponding to $a$ is \emph{aligned} to $a$ in the process model. Transitions in the original process model are represented as model moves $(\gg, a) \in T^{MM}$ and transitions in the original trace model as log moves $(a, \gg) \in T^{LM}$~\cite{carmonaConformanceCheckingRelating2018}.

An alignment is a \emph{sequence of transition firings} in the synchronous product net. 
There exists at least one alignment, such that contains only model moves, followed by log moves. However, the goal is to find the \emph{optimal} alignment between the process and trace given some cost function $c(t)$ for each transition $t$.

\begin{definition}[Cost function, Standard cost function]
A cost function $c : T \rightarrow \mathbb{R}^{+}$ is one associating a non-negative cost to each transition of the synchronous product net, incurred when firing that transition~\cite{bloemenMaximizingSynchronizationAligning2018}, such that:

$$ c(t)=\left\{
\begin{array}{ll}
0,\ t = (\gg,\tau) & (model \ move \ on \ \tau) \\
0,\  t\in T^{SM} & (synchronous\ move,\  e.g.\ (a,a)) \\
1,\ t\in T^{LM} & (log\ move,\ e.g.\ (a,\gg)) \\
1,\ t\in T^{MM} & (model\ move,\  e.g.\ (\gg,a))
\end{array}
\right.
$$
\end{definition}
Given this, the problem of finding an \emph{optimal alignment} is reduced to searching at least one execution sequence of the synchronous product with the minimum cost. 
The standard cost function ensures an optimal number of synchronous moves, i.e.\ events and process transitions that align together at the same time.

\section{Conformance checking over probabilistic events}\label{sec:algorithm}

This section presents our alignment algorithm, named \texttt{ProbCost}. 
Traditional alignment-based conformance checking only considers traces of \emph{deterministic} events. 
In our work, we instead consider a trace of $m$ events $\langle e_{0},..., e_{m-1}\rangle$ each of which can correspond to one of $n$ possible activities $\{ a_{0}, ... , a_{n-1}\}$ with the \emph{categorical probability} $p_{i,j} = p(a_j | e_i)$ that event $e_i$ is an observation of activity $a_j$. This can be modelled by a probability matrix as follows:%

\begin{equation}
\label{eq:prob matrix}
\begin{blockarray}{c@{\hspace{1pt}}ccccc@{\hspace{3pt}}}
         & e_{0} & e_{1} & ... & e_{m-2} & e_{m-1} \\
        \begin{block}{c@{\hspace{5pt}}|@{\hspace{2pt}}
    |@{\hspace{2pt}}ccccc@{\hspace{2pt}}|@{\hspace{2pt}}|}
  a_{0} & p_{0,0} & p_{1,0} & ... & p_{m-2,0} & p_{m-1,0} \\
  a_{1} & p_{0,1} & p_{1,1} & ... & p_{m-2,1} & p_{m-1,1} \\
  . & . & . & ... & . & . \\
  a_{n-1} & p_{0,n-1} & p_{1,n-1} & ... & p_{m-2,n-1} & p_{m-1,n-1} \\
\end{block}
\end{blockarray}
\end{equation}

\noindent where each row corresponds to an activity $a_j$, each column to an event $e_i$, and entries $p_{i,j}$ represent the probability that $e_i$ corresponds to $a_j$. Note that, for each event $e_{i}$, the sum of the probabilities should be 1, i.e.\ $\sum_{j=0}^{n-1}p_{i,j} = 1$.

An example trace of 3 events $\langle e_{0}, e_{1}, e_{2}\rangle$ each of which may correspond to 3 activity classes $a,b,c$ can modelled by the following probability matrix:
\begin{equation}
\label{eq:example prob matrix}
    \mathbf{P} = 
        \begin{blockarray}{c@{\hspace{1pt}}rrr@{\hspace{3pt}}}
         & e_{0}   & e_{1}   & e_{2} \\
        \begin{block}{r@{\hspace{5pt}}|@{\hspace{2pt}}
    |@{\hspace{2pt}}rrr@{\hspace{2pt}}|@{\hspace{2pt}}|}
        a & 0.3 & 0 & 0 \\
        b & 0.7 & 0.7 & 0.3 \\
        c & 0 & 0.3 & 0.7  \\
        \end{block}
    \end{blockarray}
\end{equation}
where the first event $e_{0}$ has a $0.3$ probability to be associated with activity $a$ and $0.7$ probability to be associated with $b$. 

This probability matrix can be modelled as a \emph{weighted trace model} as follows:

\begin{definition}[Weighted trace model]
Let $A$ be a set of $n$ activities and $\sigma$ a trace of $m$ events, where each $e_{i}\in\sigma$ corresponds to each $a_j\in A$ with probability $p_{i,j}$. 
A weighted trace model is a Petri net $(P^{wt},T^{wt},F^{wt},  \alpha^{wt}, w^{wt})$, where:
\begin{itemize}
    \item $P^{wt} = \{P_{0},...,P_{m}\}$ a set of $m$ places,
    \item $T^{wt} = \bigcup_{i=0}^{m-1}\bigcup_{j=0}^{n-1} \{t_{i,j} \mid p_{i,j} > 0\}$, where $t_{i,j}$ is a transition for event $e_i$ if activity $a_j$ corresponds to it with probability $p_{i,j} >0$,
    \item $F^{wt} = \bigcup_{i,j} \{(P_{j-1},t_{i,j}),\ (t_{i,j}, P_{j}) \mid t_{i,j} \in T^{wt} \}$,
    \item $\alpha^{wt}(t_{i,j}) = a_j$, for each transition $t_{i,j} \in T^{wt}$,
    \item $w^{wt}:T^{wt} \rightarrow [0,1]$, where $w^{wt}(t_{i,j}) = \{p_{i,j} \mid 0 \leq i \leq m, 0 \leq j \leq n \}$ is a function assigning a weight to each transition according to the probabilistic matrix of the event log,

\end{itemize}
\end{definition}

\begin{figure}[t]
    \centering
    \includegraphics[width=0.5\columnwidth]{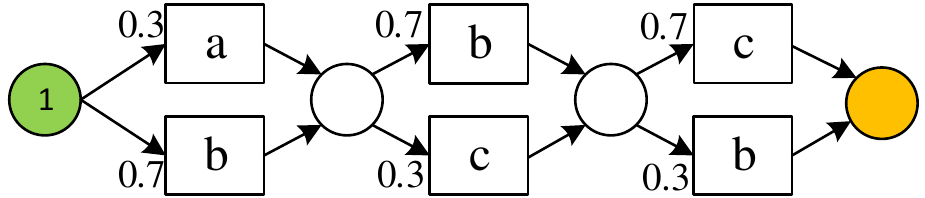}
    \caption{Weighted trace model of probability matrix~\eqref{eq:example prob matrix}.}
    \label{fig:probabilistic trace model}
\end{figure}

Note that, in contrast to deterministic trace models, a weighted trace model has multiple transitions for every event, one for each possible activity.
The weighted trace model corresponding to probabilistic matrix~\eqref{eq:example prob matrix} is shown in Fig.~\ref{fig:probabilistic trace model}. 

\begin{definition}[Weighted synchronous product net]
\label{def:synchronous product net}
Let $A$ be a set of $n$ activities, $N_{p} = (P^{p},T^{p},F^{p}, \alpha^{p})$ a process model and $\sigma$ an observed probabilistic trace of $m$ events with probability distribution $p_{i,j}$ over the activities, with the corresponding weighted trace model $N_{wt} = (P^{wt},T^{wt},F^{wt}, \alpha^{wt}, w^{wt})$. A synchronous product net is a Petri net $N_{ws} = (P^{ws},T^{ws},F^{ws}, \alpha^{ws}, w^{ws})$, such that:
\begin{itemize}
    \item $P^{ws} = P^{p} \cup P^{wt}$,
    \item $T^{ws} = (T^{MM} \cup T^{LM} \cup T^{SM})$, where 
    \begin{itemize}
        \item $T^{MM} = \{\gg\} \times T^{p} $ denotes moves on model,
        \item $T^{LM} = T^{wt} \times \{\gg\} $ denotes moves on log,
        \item $T^{SM} = \{(t_{1},t_{2}) \in T^{p} \times T^{wt} \mid \alpha^{p}(t_{1}) =  \alpha^{wt}(t_{2})\}$ denotes synchronous moves.
    \end{itemize}
    \item $F^{ws} = \{(p,(t_{1},t_{2})) \in P^{ws} \times T^{ws} \mid (p,t_{1}) \in F^{p} \vee (p,t_{2}) \in F^{wt}\} \cup \{((t_{1},t_{2}),p) \in T^{ws} \times P^{ws} \mid (t_{1},p) \in F^{p} \vee (t_{2},p) \in F^{wt}\}$,
    \item $\alpha^{ws}((t_{1},t_{2})) = (l_{1},l_{2})$ for all transitions $(t_{1}, t_{2}) \in T^{ws} $, where $l_{1} = \alpha ^{p}$ if $t_{1} \in T^{p}$, otherwise $l_{1} = \gg$; and $l_{2} = \alpha ^{wt}$ if $t_{2} \in T^{wt}$, otherwise $l_{2} = \gg$,
    \item $ w^{ws}:T^{ws} \rightarrow [0,1] $ a weight function for each transition, where $w^{ws}((t_1,t_2)) = w^{wt}(t_1)$ if $(t_1,t_2) \in (T^{LM} \cup T^{SM})$, otherwise $w^{ws}((t_1,t_2)) = 1$, 

\end{itemize}
\end{definition}

Note that the worst-case of the number of synchronous transitions is exponential to the number of events, which occurs when (i) the number of events in the trace is the same as the number of activities in the process model, and  (ii) each event has multiple transitions, and (iii) all transitions match to a corresponding labelled activity in the process model. However, the worst-case is unrealistic, because it assumes that all events are uncertain and overgeneralised enabling match every activity in the process model. 

\begin{figure}[t]
    \centering
    \includegraphics[width=0.9\columnwidth]{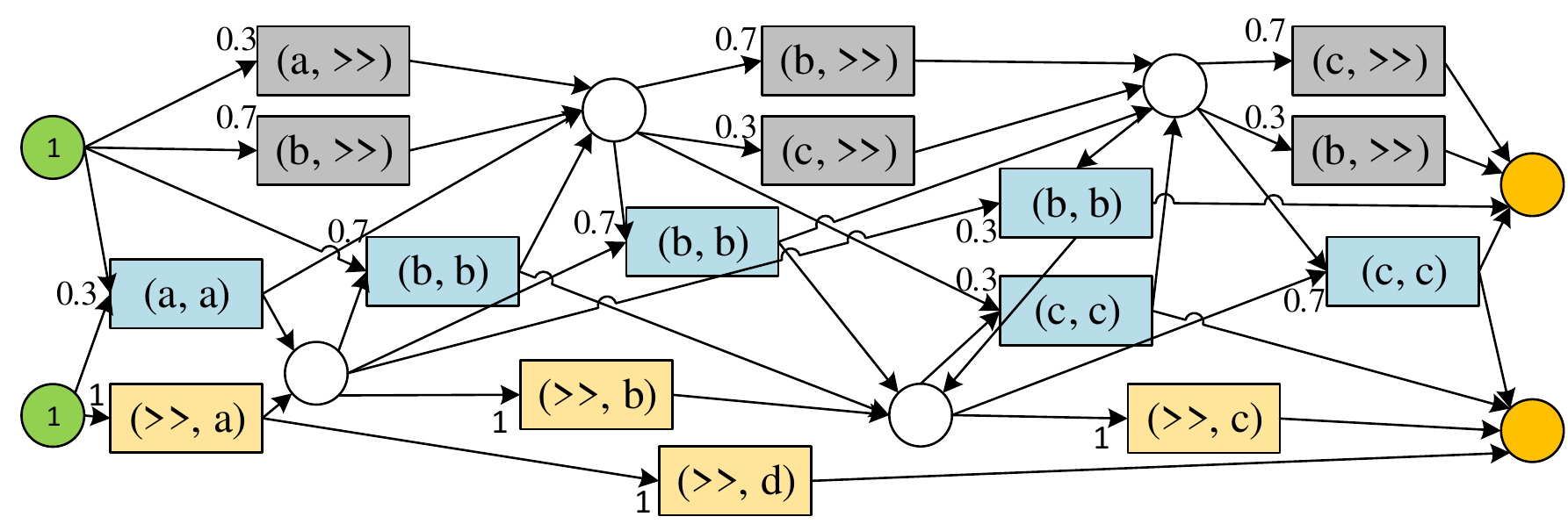}
    \caption{Synchronous product net of our example models from Fig.~\ref{fig:process model} and \ref{fig:probabilistic trace model}, annotated using weight function $w^{ws}$ and marked different colours for model moves (yellow), log moves (grey) and synchronous moves (blue).}
    \label{fig:prob_sync}
\end{figure}

Fig.~\ref{fig:prob_sync} shows the weighted synchronous product net of our example process model (Fig.~\ref{fig:process model}) and weighted trace model (Fig.~\ref{fig:probabilistic trace model}).

Knowing the probabilities of each activity given an observed event allows us to calculate the cost of each transition based on the likelihood that the corresponding activity actually occurred. This likelihood is reflected in the weight $w^{ws}$ of each transition. 
Activities with higher probability should have lower cost so that the algorithm is more likely to select them in the optimal alignment. We use a $log$ transformation to transform probability products to sums of log probabilities. Based on this, we define a new, weighted cost function $c(t)$:

\begin{definition}[Weighted cost function]
\label{def:probabilistic cost function}
Given a weighted synchronous product net $N_{ws} = (P^{ws},T^{ws},F^{ws},  \alpha^{ws}, w^{ws})$, we define a weighted cost function for each transition $t \in T^{ws}$ of the synchronous product net as follows:
$$
c(t)=\left\{
    \begin{array}{ll}
    0,\ & t = (\gg,\tau) \ (model \ move \ on \ \tau) \\
    -log(w^{ws}(t)) , & t \in T^{SM} \  \\
    -log(w^{ws}(t)) - log(\epsilon), & t \in T^{LM} \  \\
    -log(\epsilon), & t \in T^{MM} \ \\
    \end{array}
    \right.
$$

\noindent where $\epsilon \in (0, 1)$ is a parameter representing the level of confidence in the event log (see Section~\ref{sec:threshold}).
\end{definition}

Using this cost similarly to the standard deterministic alignment algorithm~\cite{adriansyahConformanceCheckingUsing2011}, we find an optimal alignment by using A* algorithm.

\textbf{To summarise}, our algorithm, \texttt{ProbCost}, includes following steps: (i) build process model and weighted trace model, (ii) construct weighted synchronous product net, (iii) use A* to find optimal alignment based on weighted cost function, (iv) tune threshold parameter $\epsilon$ based on context, (v) achieve final alignment based on tuned $\epsilon$.

\begin{example}\label{examp: first example}

Consider the process model in Fig.~\ref{fig:process model} and the example probabilistic trace in \eqref{eq:example prob matrix}, and their composed weighted synchronous product net in Fig.\ref{fig:prob_sync}, we could get the cost of the synchronous move, e.g., $(a,a)$ is $-log(0.3)$, the cost of the model move, e.g. $(\gg,a)$ is $-log(\epsilon)$, and the cost of the log move, e.g. $(a,\gg)$ is $-log(0.3)-log(\epsilon)$. The optimal alignment computed by \texttt{ProbCost} ($\epsilon=0.4$) is shown in Table~\ref{tab:prob example alignment}. Compared to the results of standard (deterministic) alignment algorithm (shown in Table~\ref{tab:standard example alignment}) which assumes the \textit{most likely} activities for each event, i.e. the trace $\langle b,b,c \rangle$, \texttt{ProbCost} chooses activity $a$ for the first event though is has a lower probability (0.3) than b (0.7) and forms a perfect alignment with process model. This is  because the cost of choosing the same trace as the standard alignment algorithm, i,e, $\langle b,b,c \rangle$ is higher than the cost of trace $\langle a,b,c \rangle$. \looseness=-1

This fits our intuition that even though $a$ is less likely to have occurred given our observation of the first event, the process model captures the knowledge that $a$ is \emph{expected} to occur (e.g.\ based on prior observations or physical restrictions), thus increasing our perceived likelihood for $a$. In other words, we use our confidence in the modelled process to mitigate the uncertainty of the event log.

\end{example}

\begin{table}[!t]%
\caption{Alignment results of 3 events with probability matrix~\eqref{eq:example prob matrix} and the process model from Fig.~\ref{fig:process model}.}%
\label{tab:example alignment}%
\begin{subtable}{.5\linewidth}%
\caption{Standard alignment}%
\label{tab:standard example alignment}
\centering

\begin{tabular}{c|c|c|c|c|}
Event log     & $\gg$ & $b$ & $b$ & $c$  \\ \hline
Process model & $a$ & $\gg$ & $b$ & $c$ 
\end{tabular}

\end{subtable}%
\begin{subtable}{.5\linewidth}
\caption{Probabilistic alignment}
\label{tab:prob example alignment}
\centering
\begin{tabular}{c|c|c|c|}
Event log     & $a$ & $b$ & $c$  \\ \hline
Process model & $a$ & $b$ & $c$ 
\end{tabular}
\end{subtable}

\end{table}

\begin{example}
Consider again the same process model and event log with Example~\ref{examp: first example}, we change the $\epsilon$ in cost function as $0.8$. \texttt{ProbCost} gets the same optimal alignment with the standard alignment algorithm, as shown in Table~\ref{tab:standard example alignment}. 

Compared to the result of Example~\ref{examp: first example}, it shows that the $\epsilon$ in our algorithm is able to control the acceptable probability of events, i.e., instead of blindly accept the \textit{expected activity} by the process model, the $\epsilon$ is able to determine whether we trust the event log's indication of activity or the process model's expectation of activity. We illustrate the $\epsilon$ formally in Section~\ref{sec:threshold}.

\end{example}

\noindent\textbf{Complexity analysis} Compared to the standard A* alignment approach whose worst-case space complexity is linear to the number of reachable state of the synchronous product net~\cite{carmonaConformanceCheckingRelating2018}, we have the same number of reachable state, therefore, we have same space complexity. Furthermore, time complexity depends on the number of transitions between initial and final states of the synchronous product net~\cite{carmonaConformanceCheckingRelating2018}. Therefore, the worst-case time complexity of \texttt{ProbCost} can be exponential compared to the standard approach, when the worst-case number of transitions occurs in the synchronous product net. Furthermore, an empirical evaluation of computation time is discussed in Section~\ref{sec:deviation results}.

\section{Threshold parameter $\epsilon$}\label{sec:threshold}
Our weighted cost function $c(t)$ includes a ``threshold'' parameter $\epsilon$. As hinted in the previous section, $\epsilon$ allows the user to control the level of confidence, or \emph{trust}, in the modelled process given an uncertain event log. An intuitive way to express this is by posing the following question:
\begin{quote}
    \emph{How probable does an event need to be for it to move synchronously with a matching activity in the process model?}
\end{quote}

Let us explore a minimal example whereby $\epsilon$ allows us to control the answer to that question.

Consider a trace with a single event $e_{0}$ that corresponds to either of two possible activities $a$ and $b$ with probabilities $p_{0,a} = x$ and $p_{0,b} = 1-x$, for some $x\in [0,1]$. We want to check conformance of this trace with a model containing $a$ as a single step in the process. In this case, we have 2 possible optimal alignments shown in Table~\ref{tab:threshold_e1}. 

\begin{table}[t]
    \caption{Possible optimal alignments of a single event with 2 possible activities $a$, $b$ and process model with single step $a$.}
    \label{tab:threshold_e1}
    \begin{subtable}{.5\linewidth}
      \centering
        \caption{Assuming $a$ occurred.}
        \label{tab:threshold_e1_1}

        \begin{tabular}{c|c|}
            Event log     & $a$   \\ \hline
            Process model & $a$  
        \end{tabular}
    \end{subtable}%
    \begin{subtable}{.5\linewidth}
      \centering
        \caption{Assuming $b$ occurred.}
        \label{tab:threshold_e1_2}
        \begin{tabular}{c|c|c|}
            Event log     & $b$ & $\gg$  \\ \hline
            Process model & $\gg$ & $a$ 
        \end{tabular}
    \end{subtable} 
\end{table}

The first (Table~\ref{tab:threshold_e1_1}) contains a synchronous move $a$, which assumes that the event corresponds to activity $a$. The second (Table~\ref{tab:threshold_e1_2}) considers that the event corresponds to activity $b$, leading to a model move and a log move. The optimal choice between the two should be based on the probabilities of $a$ and $b$. The question is \emph{what value does $x$ need to be for the first alignment to be optimal?}

In the extreme cases, if $x=1$ then $a$ occurred with $100\%$ probability, so the first alignment should be optimal, and symmetrically if $x=0$ then the second alignment should be optimal. Furthermore, if we only consider the most likely activity as the one that actually occurred, which is in fact standard in activity recognition~\cite{sztylerSelftrackingReloadedApplying2016}, then the second alignment would be optimal only when $x < 0.5$.

However, we posit that this decision may be influenced by the process model. For instance, consider the extreme case where the process model reflects domain knowledge that \emph{only activity $a$ is actually possible}, whereas activity $b$ is not possible \emph{at all}\footnote{Of course, such an extreme model would defeat the purpose of conformance checking to begin with, since we already know what is possible, but it helps illustrate our point.}. A high probability $p_{0,b}$ might then be attributed, for example, to noise in the data, and the second alignment should \emph{always} be sub-optimal. 

More generally, we may choose to \emph{trust} the reality (or our expectation of it) as reflected by the process model more than a noisy and uncertain event log, i.e.\ perform a synchronous move on $a$ even when $x < 0.5$.

In our approach, the threshold parameter $\epsilon$ allows us to control how low $x$ can be for $a$ to be chosen by the algorithm. Given Definition~\ref{def:probabilistic cost function}, the total cost of the alignment in Table~\ref{tab:threshold_e1_1} is $-log(x)+log(\epsilon)$ (synchronous move on $a$), whereas the total cost of the alignment in Table~\ref{tab:threshold_e1_2} is $-log(1-x)-log(\epsilon)$ (move on log $b$ and move on model). The former is optimal under the condition: %
\begin{equation}
\nonumber
\frac{x}{1-x} > \epsilon^{2} \label{eq:linear threshold}
\end{equation}%
For the synchronous move on the model's expected activity $a$ to be optimal (i.e.\ to assume that $a$ occurred) the ratio of $p_{0,a}$ to $p_{0,b}$ should be more than $\epsilon^2$. 

\begin{figure}[!tbp]
  \centering
  \begin{minipage}[b]{0.49\textwidth}
    \includegraphics[width=\textwidth]{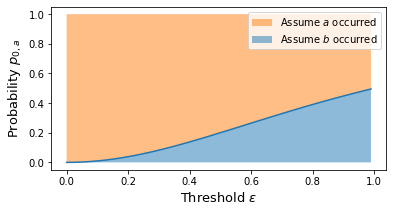}
    \caption{Relationship between $\epsilon$, $p_{0,a}$, and optimal alignment.}
    \label{fig:threshold}
  \end{minipage}
  \hfill
  \begin{minipage}[b]{0.49\textwidth}
    \includegraphics[width=\textwidth]{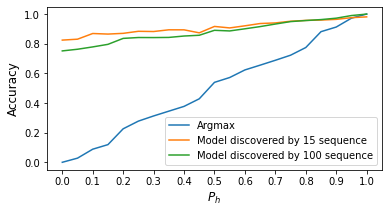}
    \caption{Trace recovery accuracy for different number of traces in model discovery.}
    \label{fig:trace recovery}
  \end{minipage}
\end{figure}

The relationship between $p_{0,a}$, $\epsilon$ and the assumed activity is shown in Fig.~\ref{fig:threshold}. For each combination of values for $p_{0,a}$ and $\epsilon$ above the line, the algorithm assumes $a$ occurred and picks the first alignment as optimal. Otherwise below the line, it assumes $b$ occurred and picks the second alignment as optimal. The lower the threshold, the lower the probability of $a$ needs to be for a synchronous move, and therefore the more we trust the process model instead of the probability distribution of the event log. With $\epsilon \in (0,1)$, the threshold for the probability of $a$ can be set to be anywhere in $(0, 0.5)$.

It is obvious that if we consider more complex process models and larger traces, with multiple possible activities per event, the interaction between $\epsilon$, the activity probabilities and the alignment costs become harder to analyse. This makes the selection of the appropriate value of $\epsilon$ more difficult.
However, a general principle applies: \emph{when $\epsilon$ is closer to $1$, the algorithm trusts the probabilities of the event log and considers the most probable activities. As the value of $\epsilon$ decreases, the algorithm puts more trust in the process model and accepts activities with lower probability in the event log if they align to those in the model.}

This selection of the appropriate value for $\epsilon$ depends on the context of each case study and our confidence in the process model. In many of our examples when choosing between possible optimal alignments, we have observed $\epsilon$ to be the threshold for the \emph{ratio of probabilities} of the activities corresponding to each alignment. This pattern is also observed in our experimental study. Pending further research on the role of $\epsilon$ and methods to determine the appropriate value in each case, we select its value based on (i) the developed intuition (higher values mean higher trust on the log), (ii) the ratio of the involved probabilities as a rule of thumb, and (iii) empirical exploration by training on the training dataset. \looseness=-1

\section{Experimental study}\label{sec:experiment}

We first evaluate \texttt{ProbCost}'s ability to replicate the same trace recovery results of Bogdanov et al.~\cite{bogdanovConformanceCheckingStochastically2022a,bogdanovTraceRecoveryStochastically2022} and validate our method's subsumption of their approach as a baseline. Next, we assess whether \texttt{ProbCost} has outstanding performance in deviation detection compared to the standard (deterministic) alignment algorithm and Bogdanov et al.'s approach.

Our experiment employs three publicly available real-world datasets: (i) BPI Challenge (BPIC) 2012\footnote{https://www.win.tue.nl/bpi/doku.php?id=2012:challenge} related to personal loan processes, which contains 13087 cases over 36 activities, e.g., assessing the application, calling after sent offers, etc., (ii) BPIC 2019\footnote{https://icpmconference.org/2019/icpm-2019/contests-challenges/bpi-challenge-2019/} related to purchase order handling processes including 251734 cases over 42 activities, e.g., submit order, payment, etc., and (iii) ADLs daily log~\cite{sztylerSelftrackingReloadedApplying2016}, which records individuals' daily routines for 10 days, including 13 types of activities, e.g., eating, take medicine, etc. Both BPIC datasets have been used to evaluate algorithms in uncertain data settings~\cite{bogdanovConformanceCheckingStochastically2022a,pegoraroEfficientConstructionBehavior2020}.

\subsection{Evaluating trace recovery}\label{sec:trace recovery}

In our first experiment, we investigate the ability of \texttt{ProbCost} to recover the real trace from probabilistic event log (trace recovery) under the same assumptions as Bogdanov et al. 
Replicating their strategy and use the same BPIC 2012 dataset, we first discover a process model based on randomly selected traces. We then generate probabilistic events by adding noise to the original activities of another 100 traces that are not used in discovery. 

For each event in each of the traces we attach a second alternative activity to its original label, selected randomly from the set of activities in the dataset. The original activity is assigned a randomly chosen probability $p$, whereas the added activity is assigned $1-p$. A parameter $P_{h} \in [0,1]$ specifies the proportion of the original activities that will be assigned higher probabilities than their added alternatives ($p > 1-p$). When $P_{h}=1$ all the original activities are assigned higher probability than the added alternatives, and vice versa when $P_{h}=0$. 

We then perform conformance checking with our algorithm, and extract the optimally aligned sequences of events as the recovered traces. We calculate the recovery accuracy by dividing the number of correctly recovered events (compared to the original ones) by the total number of events, and compare the results against the Argmax sequence of most probable activities for the events.

The threshold parameter $\epsilon$ is set to a low value (0.01) to simulate the intention of Bogdanov et al.\ to maximally align activities to the process model regardless of probability. For the model discovery we use the Inductive Miner ~\cite{adriansyahConformanceCheckingUsing2011} as implemented in PM4Py~\cite{PM4PYpaper2019}, on separate sets of 15 and 100 traces. Our results compared to Argmax are shown in Fig.~\ref{fig:trace recovery},

Our results seem to match those of Bogdanov et al~\cite[Figure 5]{bogdanovTraceRecoveryStochastically2022}. 
Specifically, when $P_{h}$ is 0, recovery accuracy based on the model discovered by 15 traces (0.81) is better than the result using 100 traces (0.75), which is same to their result. As $P_{h}$ increases, the accuracy increases, until $P_{h}$ reaches 1, where the same result as Argmax is obtained. This validates that their approach can be replicated by our algorithm by setting a low value of $\epsilon$, close to 0 (e.g.\ 0.01) as a baseline. \looseness=-1

\subsection{Evaluating deviation detection}\label{sec:eva deviation detection}

\subsubsection{Experiment design}
In this experiment, we evaluate the performance of our algorithm when detecting deviations on all three datasets, i.e.\ identifying events that do not fit the process model. In the context of alignment-based conformance checking, a log move indicates a deviation (the event is inconsistent with the process model), while a synchronous move means a normal occurrence (the event fits with the activity in the process model). We compare the performance of our approach with the standard alignment algorithm over the Argmax sequences and the approach of Bogdanov et al. Note that in contrast to the assumption we made in Section~\ref{sec:trace recovery}, even if events fit the process model but occur with low probability, we still regard them as deviations.

We aim to synthesise a set of probabilistic event traces in such a way that we can control the ratio of deviations, and there exist low probability activities that are not deviations. We accomplish this as follows.

For the two BPI datasets, we first perform model discovery using 20 randomly chosen traces, and choose another 100 traces from the rest to generate probabilistic events using the same setting as previously. For the smaller ADL dataset which contains 10 traces of daily logs, we randomly chose 5 traces for model discovery and the remaining 5 for generating probabilistic events. We use $P_{h} = 0$ so that all the added activities have higher probabilities than the original ones ($p<1-p$).

Standard conformance checking over the traces against the discovered model yielded 100\%, 96\% and 100\% fitness for the three datasets respectively. However, for the sequences of \emph{added} activities, the average fitness and standard deviation (SD) are 23\% (0.08 SD), 8\% (0.11 SD), and 49\% (0.15 SD), respectively. This means that the original traces conform to the process model (even though they were not used in discovery), whereas the added activities do not have a good fit. Based on this, we define deviations in a probabilistic trace by considering the original activities as \emph{normal occurrences} and the added ones as \emph{deviations}. 

Furthermore, under our assumption of noisy data, we introduce a \emph{deviation confidence} parameter $T_{d}$. Specifically, for each event, if
the odds of the original activity happening, i.e.\ the ratio of the probability of the original activity $p$ to that of the alternative is higher than $T_{d}$, i.e.\ $\frac{p}{1-p} \geq T_{d}$, then we classify this event as a normal occurrence. Otherwise, the event is a deviation.

For example, for $T_{d} = 0.5$, if $p < 0.333$ then the alternative occurred (true deviation). However, if $p=0.4$, then we determine that the event is in fact an actual occurrence. Note that $p < 0.5$ always, since $P_{h} = 0$. 

If $T_{d}$ is close to 0, the original activity is a normal occurrence even when its probability is very low, as long as it is non-zero, resulting in very few deviations. As $T_d$ increases from 0 to 1, the proportion of the two classes changes, i.e., the number of normal events decreases, while the number of deviations increases.

Therefore, in this setup we control the ratio of deviations using $T_d$ and normal occurrences have low probabilities ($T_h = 0$).

We use accuracy, F1-score, sensitivity ($\frac{TP}{TP+FN}$) and specificity ($\frac{TN}{TN+FP}$) to measure the performance of deviation detection based on the counts of True Positives (TP), False Positives (FP), True Negatives (TN) and False Negatives (FN)~\cite{arifogluDetectionAbnormalBehaviour2019}. We also use Geometric mean (G-mean, i.e., $\sqrt{sensitivity \times specificity}$) metric to balance both sensitivity and specificity on imbalanced datasets~\cite{kubat1997addressing}.

\subsubsection{Tuning threshold $\epsilon$} 
\begin{figure}[t]
    \centering
    \includegraphics[width=\columnwidth]{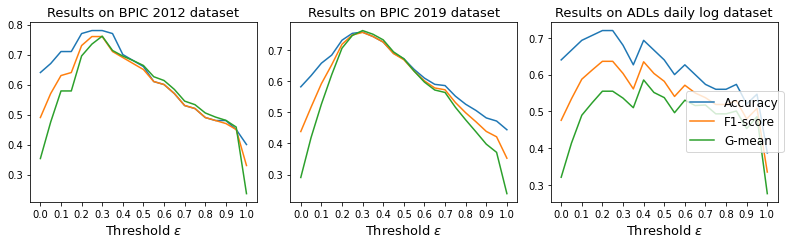}
    \caption{Deviation detection performance for different values of $\epsilon$ ($T_{d}=0.25$).}
    \label{fig:different confidence threshold}
\end{figure}

Next, we set the \textit{deviation confidence} $T_{d}$ to 0.25, which means the probability of the original activity should be higher than 0.2 to be considered as a normal occurrence. We evaluate the performance of \texttt{ProbCost} over the development set (70\% of the whole data) under different values of $\epsilon$, ranging from 0.05 to 1 with a 0.05 step. The accuracy, F1-score and G-mean are calculated for each value of $\epsilon$ to determine the optimal value that yields the best algorithm performance, as these metrics are suitable for evaluating imbalanced datasets. The results are presented in Fig.~\ref{fig:different confidence threshold}. The best performance was achieved for $\epsilon$ between $0.25$ to $0.3$ for all three datasets (variation due to classes proportions), which is basically equal to the deviation confidence $T_{d}$. This follows the intuition and empirical results discussed in Section~\ref{sec:threshold}. Based on this, we set the $\epsilon$ as the value of $T_{d}$ for further experiments.

\subsubsection{Results}\label{sec:deviation results}

\begin{figure}[t]
    \centering
    \includegraphics[width=\columnwidth]{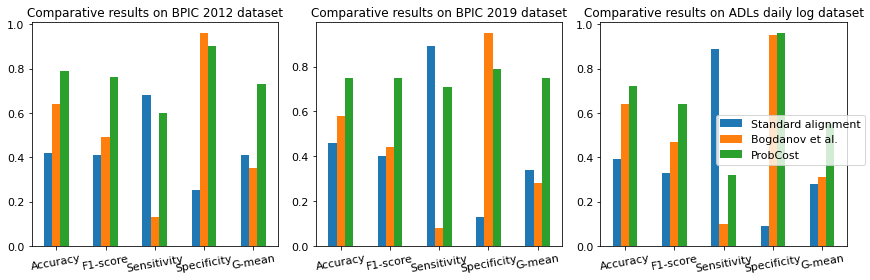}
    \caption{Results for deviation detection on different datasets ($T_{d}=0.25$).}
    \label{fig:performance comparision}
\end{figure}

We assess the deviation detection results using our algorithm ($\epsilon=0.25$) compared to the standard alignment conformance checking algorithm as implemented in PM4Py and the approach by Bogdanov et al. as replicated by our algorithm with $\epsilon=0.01$. The comparative results are shown in Fig.~\ref{fig:performance comparision}.

Based on these results, \texttt{ProbCost} performs better than both the standard alignment algorithm and Bogdanov et al in \textit{Accuracy}, \textit{F1-score} and \textit{G-mean} metrics. The last one has the lowest \textit{sensitivity} and mostly highest \textit{specificity} because it always selects the activities that fit in the process model even when their probability is very low, leading to more false negatives. In contrast, the standard alignment algorithm has highest \textit{sensitivity} and lowest \textit{specificity}, because it may ignore the normally occurring activity even if its probability is close to the probability of the alternative, leading to more false positives. \texttt{ProbCost} achieves better \textit{G-mean} score than the other two approaches, it suggests that it is better at correctly identifying both positive and negative cases, resulting in a better overall performance. In brief, our algorithm can better balance the confidence between the process model and the uncertain event log to obtain better results than both the others.

We further investigate the three algorithms under different deviation confidence values $T_{d}$. For this, we iterate $T_{d}$ from 0 to 1 with 0.05 step to generate deviation data before running the 3 algorithms. 
As mentioned previously, we set $\epsilon=T_d$. The approach of Bogdanov et al.\ is replicated by our algorithm with $\epsilon=0.01$. \looseness=-1

\begin{figure}[t]
    \centering
    \includegraphics[width=\columnwidth]{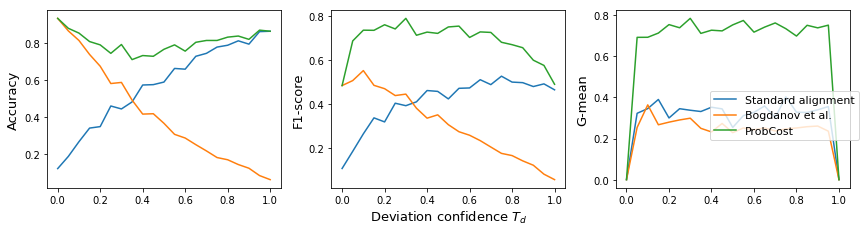}
    \caption{BPIC 2019 results for different values of $T_{d}$.}
    \label{fig:comparing3algorithms}
\end{figure}

\texttt{ProbCost} outperforms the others for all the values of $T_{d}$ in three datasets. Results for different datasets show similar tendencies, we only show the results of BPIC 2019 in Fig.~\ref{fig:comparing3algorithms} due to space considerations.
For lower values, as expected, our performance matches that of Bogdanov et al.\ as no matter the probability of the original activity, it can always be aligned with the expected activity in the process model to identify a normal occurrence. However, the standard alignment algorithm gives the lowest accuracy and F1-score because it chooses the most probable activity for each event, resulting in more false positives. All three algorithms produce the lowest G-mean when $T_d=0$, because there are no deviations, causing the sensitivity to be 0. Conversely, there are no normal occurrences when $T_d=1$, causing specificity to be 0. 

As the deviation confidence $T_d$ increases, the accuracy and F1-score of approach by Bogdanov et al.\ decrease, while they increase for the standard algorithm. This happens because as $T_d$ increases, the original activities need to have higher probability to be considered as normal occurrences, but Bogdanov et al.\ accept activities expected by the process model but with little probability, resulting in more false negatives. 
When $T_d$ is increased to 1, our approach yields the same accuracy and F1-score as the standard alignment algorithm. As we put more trust on the event log (higher $\epsilon$), we end up considering the most probable activities, similarly to the standard alignment algorithm.

Finally, we assess the time performance of \texttt{ProbCost} on three datasets with different sizes of events and process models. The results (shown in Table~\ref{Tab:execution time}) indicate that the execution time of \texttt{ProbCost} is longer than the standard approach as we discussed in complexity analysis (Section~\ref{sec:algorithm}), because the number of transitions increases, but it still within a reasonable range. It demonstrates that our approach is applicable in practice.

\begin{table}[t]

\caption{Comparison of computation time between standard alignment and \texttt{ProbCost} for different datasets.}
\resizebox{\columnwidth}{!}
{
\begin{tabular}{c|c|cccc|cc|cc}
\hline
\multirow{2}{*}{Datasets} & \multirow{2}{*}{\#Cases} & \multicolumn{4}{c|}{\#Events per case} & \multicolumn{2}{c|}{Process model} & \multicolumn{2}{c}{\begin{tabular}[c]{@{}c@{}}Computation time \\ per case (in seconds)\end{tabular}} \\ \cline{3-10} 
                          &                          & min     & max    & avg     & median    & \#places      & \#transitions      & Standard                                               & \texttt{ProbCost}                                                 \\ \hline
BPIC 2012                 & 100                      & 3       & 56     & 24.7    & 27.5      & 44            & 59                 & 356.150                                                & 865.201                                              \\
BPIC 2019                 & 100                      & 1       & 10     & 5       & 5         & 12            & 18                 & 0.174                                                  & 0.629                                                \\
ADLs daily log            & 5                        & 7       & 22     & 15      & 14        & 26            & 41                 & 48.094                                                 & 93.322                                               \\ \hline
\end{tabular}
}
\label{Tab:execution time}
\end{table}

Overall, our approach demonstrates more resilience to noise and uncertainty in the events, and flexible control of how much the algorithm should trust the event log over the structure of the process model via $\epsilon$.

\section{Conclusion}\label{sec:conclusion}

In this paper, an extended alignment conformance checking algorithm is proposed for use with probabilistic event logs. Specifically, we propose a weighted trace model for events with a categorical probability distribution, a weighted alignment cost function, and a custom threshold parameter that controls the level of confidence on the event log vs.\ the process model. Furthermore, we compared the complexity results with deterministic alignment approach. Two sets of experimental studies comparing our approach to the deterministic alignment algorithm and the recent relevant approach by Bogdanov et al.\ show that our algorithm takes into consideration occurrence probabilities explicitly and is able to accommodate activities with lower, but sufficiently high probabilities if they fit the model. Based on this, we argue that our algorithm can better tolerate noise in the event log by leveraging the knowledge captured in the process model with the aim of reducing the false positive and false negative rates in deviation detection. As such, it is better suited to perform conformance checking under an uncertain, noisy event log, such one produced by sensors or AI-based algorithms~\cite{IzackUncertainChallenge2021}. \looseness=-1

In future work, we plan to further investigate the formal results of the effect of $\epsilon$ to guide the selection of an appropriate value, as we are in favour of empirical evaluation at this stage. We also plan to improve the efficiency of algorithm and explore other types of uncertainty of event data, e.g., repeating or missing events. 

\bibliographystyle{splncs04}
\bibliography{ref}

\end{document}